\documentclass[Afour,sageh,times]{sagej}
\usepackage{moreverb,url}
\usepackage{xcolor}
\definecolor{mygray}{gray}{0.6}

\usepackage[colorlinks,bookmarksopen,bookmarksnumbered,linkcolor=blue,citecolor=blue,urlcolor=blue]{hyperref}
\usepackage{makecell}

\usepackage{bm}
\usepackage{stfloats}   % enable multi-column figures at the top and bottom 
\usepackage{tabularx} 
\usepackage{array}
\usepackage{ragged2e}
\usepackage{multirow}
\definecolor{myGray}{rgb}{0.85,0.85,0.85}
\definecolor{myBlue}{rgb}{0.5294, 0.8078, 0.92156}
\usepackage{tikz}
\usepackage{tikz-3dplot}
\usetikzlibrary{calc,patterns,angles,quotes}
\usepackage{pgfplots}
\pgfplotsset{compat=1.8}
\usepgfplotslibrary{statistics}
\usepackage{pifont} 
\newcommand{\cmark}{\ding{51}}
\newcommand{\xmark}{\ding{55}}
\usepackage{colortbl}
\usepackage{svg}
\usepackage{xurl}
\newcommand\BibTeX{{\rmfamily B\kern-.05em \textsc{i\kern-.025em b}\kern-.08em
T\kern-.1667em\lower.7ex\hbox{E}\kern-.125emX}}

\def\volumeyear{2023}
\begin{document}

\runninghead{Wenda Zhao et al.}

\title{UTIL: An Ultra-Wideband Time-Difference-of-Arrival Indoor Localization Dataset}

\author{Wenda Zhao\affilnum{1}, Abhishek Goudar\affilnum{1}, Xinyuan Qiao\affilnum{1}, and Angela P. Schoellig\affilnum{1,2}}

\affiliation{\affilnum{1}University of Toronto Institute for Aerospace Studies (UTIAS), the University of Toronto Robotics Institute, Vector Institute for Artificial Intelligence, Toronto, Canada; \affilnum{2}Technical University of Munich, Munich Institute for Robotics and Machine Intelligence (MIRMI), Munich, Germany}

\corrauth{Wenda Zhao, University of Toronto, Ontario, M3H 5T6, Canada.}

\email{wenda.zhao@robotics.utias.utoronto.ca}

\begin{abstract}
Ultra-wideband (UWB) time-difference-of-arrival (TDOA)-based localization has emerged as a promising, low-cost, and scalable indoor localization solution, which is especially suited for multi-robot applications. However, there is a lack of public datasets to study and benchmark UWB TDOA positioning technology in cluttered indoor environments. We fill in this gap by presenting a comprehensive dataset using Decawave's DWM1000 UWB modules. To characterize the UWB TDOA measurement performance under various line-of-sight (LOS) and non-line-of-sight (NLOS) conditions, we collected signal-to-noise ratio (SNR), power difference values, and raw UWB TDOA measurements during the identification experiments. We also conducted a cumulative total of around 150 minutes of real-world flight experiments on a customized quadrotor platform to benchmark the UWB TDOA localization performance for mobile robots. The quadrotor was commanded to fly with an average speed of 0.45 m/s in both obstacle-free and cluttered environments using four different UWB anchor constellations. Raw sensor data including UWB TDOA, inertial measurement unit (IMU), optical flow, time-of-flight (ToF) laser altitude, and millimeter-accurate ground truth robot poses were collected during the flights.  The dataset and development kit are available at \url{https://utiasdsl.github.io/util-uwb-dataset/}. 

\end{abstract}


\maketitle

\addtocounter{section}{0}
\section{Introduction}
% introduce UWB
Accurate and reliable indoor localization is a crucial enabling technology for many robotics applications, ranging from warehouse management to monitoring tasks. Over the last decade, ultra-wideband (UWB) radio technology has been shown to provide high-accuracy and obstacle-penetrating time-of-arrival (TOA) measurements that are robust to radio-frequency interference~\citep{zafari2019survey}. 
UWB chips have been integrated in the latest generations of consumer electronics including smartphones and smartwatches to support spatially-aware interactions~\citep{uwbNearbyInteration, QorvoUWB, AndroidUWB}. During the FIFA World Cup 2022, UWB localization technology was used, for the first time, in an official football tournament to enhance the Video Assistant Referee (VAR) system by providing reliable, low-latency, and decimeter-level accurate ball tracking information~\citep{uwbAdidas, ballSensor, ht-ball}. 

% introduce UWB TDOA: scalability
Similar to the Global Positioning System (GPS)~\citep{enge1994global}, an UWB-based positioning system requires UWB radios (also called anchors) to be pre-installed in the environment as a constellation with known positions, which in turn serve as landmarks for positioning. In  robotics~\citep{nguyen2021range, pfeiffer2021computationally}, the two common ranging schemes used for UWB localization are \textit{(i)} two-way ranging (TWR) and \textit{(ii)} time-difference-of-arrival (TDOA).
\begin{figure}[!ht]
    \centering
    \includegraphics[width=.48\textwidth]{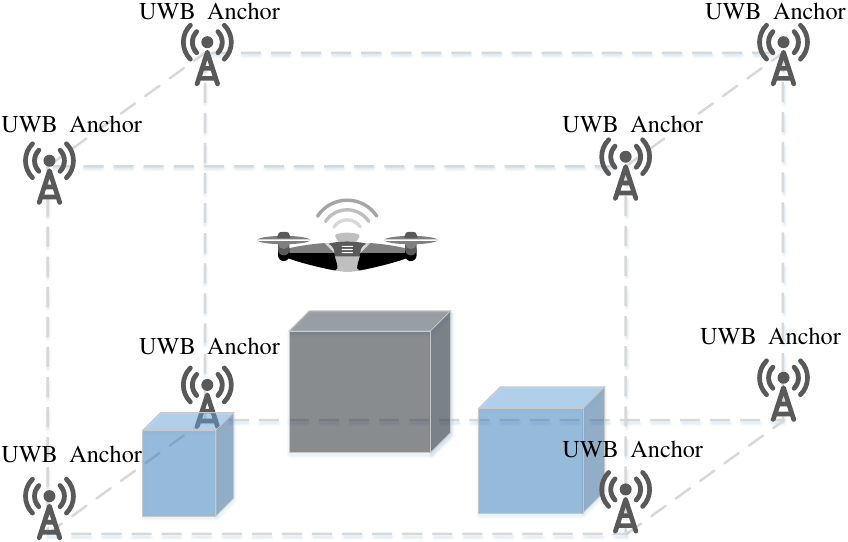}
    \caption{An UWB TDOA localization system in an indoor environment cluttered with wooden (blue boxes) and metal (the gray box) obstacles. UWB anchors are pre-installed with known positions in the space. The quadrotor, equipped with an UWB tag, receives TDOA measurements from the anchors for localization. }
    \label{fig:uwb-system}
\end{figure}
In TWR, the UWB module mounted on the robot (also called tag) communicates with an anchor and acquires range measurements through two-way communication. In TDOA, UWB tags compute the difference between the arrival times of the radio packets from two anchors as TDOA measurements. Compared with TWR, TDOA does not require active two-way communication between an anchor and a tag, thus enabling localization of a theoretically unlimited number of devices~\citep{hamer2018self}. However, UWB TDOA-based localization systems still encounter difficulties in cluttered indoor environments (see Figure~\ref{fig:uwb-system}). Delayed and degraded radio signals caused by non-line-of-sight (NLOS) and multi-path radio propagation can greatly deteriorate positioning accuracy. In order to achieve reliable UWB TDOA-based positioning in complex indoor environments, novel estimation algorithms are required to improve localization accuracy and robustness.  

% introduce the dataset: (1) content (2) contribution
% Note: compute the overall flight time: (2x3x6 + 6x3x2) * 125 =  9000 sec = 150 min
%       compute the approximate speed: 0.35 * 1.26 = ~ 0.45 m/s
To foster research in this domain, this paper presents a comprehensive UWB TDOA dataset collected in a variety of cluttered indoor environments, including different types of static and dynamic obstacles. Low-cost DWM1000 UWB modules~\citep{Datasheet} were used to construct cost-efficient indoor positioning systems for data collection. The dataset includes two parts: \textit{(i)}~an UWB TDOA identification dataset and \textit{(ii)}~a flight dataset. The goal of the identification dataset is to characterize the UWB TDOA measurement performance in line-of-sight (LOS) and non-line-of-sight (NLOS) conditions. It includes low-level UWB signal information such as signal-to-noise ratio (SNR) and power difference values. To create the NLOS scenarios, obstacles made of different materials commonly found in indoor settings were used, including cardboard, metal, wood, plastic, and foam. In the flight dataset, we conducted a cumulative total of roughly 150~minutes of real-world flights with a customized quadrotor and collected a comprehensive multimodal dataset to benchmark UWB TDOA localization performance for three-dimensional robot pose estimation. The quadrotor was commanded to fly with an average speed of $0.45$ m/s in both obstacle-free and cluttered indoor environments with static and dynamic obstacles using four different anchor constellations. Raw sensor data including UWB TDOA, inertial measurement unit (IMU), optical flow, time-of-flight (ToF) laser altitude, and millimeter-accurate ground truth robot poses data from a motion capture system were collected during the flights. 

% provide hints for data usage
The intended users of this dataset are researchers who are interested in UWB TDOA-based localization technology. The dataset can be used to model UWB TDOA measurement errors under various LOS and NLOS conditions. Also, users can study the UWB TDOA-based positioning performance \textit{(i)} under different UWB anchor constellations, \textit{(ii)} with and without obstacles, and \textit{(iii)} using the centralized and decentralized TDOA mode (introduced in Section~\ref{sec:uwb_system}). Further, the users of this dataset are encouraged to design novel and practical estimation algorithms to enhance the accuracy and robustness of UWB TDOA-based positioning in cluttered indoor environments.

% claim the contribution, 
The main contributions of this dataset are as follows:
\begin{itemize}
  \item An identification dataset for UWB TDOA measurements in a variety of LOS and NLOS scenarios involving obstacles of different materials, including cardboard, metal, wood, plastic, and foam.
  
  \item A comprehensive multimodal dataset from roughly 150~minutes of real-world flights in both obstacle-free and cluttered indoor environments, featuring both static and dynamic obstacles. We collected centralized and decentralized UWB TDOA measurements using four different anchor constellations.
 
\end{itemize}

\section{Related Work}
\label{sec:related-work}
\definecolor{Gray}{rgb}{0.85,0.85,0.85}
% note: use "m" column type to vertically center the content.
% >{\centering\arraybackslash}m{2.0cm}
\begin{table*}[!tb]
\small
  \centering
  \setlength{\tabcolsep}{7.0pt}
  \renewcommand{\arraystretch}{1.5}
  \captionsetup{width=1.0\linewidth}
  \caption{Public UWB dataset for mobile robot localization. The LOS/NLOS testing refers to experiments to identify and model UWB measurements under LOS/NLOS scenarios. Time Domain UWB modules are high-performance UWB radio sensors originally developed by Time Domain Corporation~\citep{TimeDomain}. Time Domain UWB modules can provide more accurate two-way ranging (TWR) measurements at a higher price compared to low-cost Decawave's DWM1000 modules. }
  \begin{tabularx}{\linewidth}{>{\centering\arraybackslash}m{2.0cm} c c c c c c c}
  \toprule
  \makecell{Reference}                & \makecell{UWB\\ Mode}         & \makecell{LOS\slash NLOS\\ Testing} &\makecell{Anchor\\ Constellations}  &\makecell{Static\\ Obstacle}  & \makecell{Dynamic\\ Obstacle} & Dimension      & \makecell{UWB\\ Module}\\ 
  \midrule
   \citep{bregar2018improving}   & TWR     & \cmark   & 4     & \cmark   & \xmark  & 2D  & DWM1000  \\
   \citep{li2018accurate}            & TWR     & \xmark   & 1     & \xmark   & \xmark  & 3D  & Time Domain   \\
   \citep{barral2019multi}           & TWR     & \cmark   & 1  & \cmark & \xmark    & 2D  & DWM1000   \\
   \citep{arjmandi2020benchmark}      & TWR     & \xmark   & 5  & \xmark    & \xmark & 3D   & Time Domain  \\ 
   \citep{queralta2020uwb}            & TWR     & \xmark   & 4    & \xmark   & \xmark  & 3D  &DWM1001     \\
   \citep{nguyen2021ntu}              & TWR     & \xmark   & 3    & \xmark   & \xmark  & 3D  & Time Domain   \\
   \citep{ledergerber2019ultra}       & \makecell{TWR and \\CIR values}  & \cmark & 4  & \makecell{Wood, metal, and \\ fabric (chair)} & \xmark  & 2D & DWM1000 \\

   \citep{moron2023benchmarking}      & \makecell{TWR}  & \cmark & 3  & \cmark & \xmark  & 3D & DWM1001 \\
   \citep{Industrial-UWB-23}      & \makecell{TWR}  & \cmark & 1  & \cmark & \xmark  & 3D & DWM1000 \\
   \citep{pfeiffer2021computationally}  & \makecell{TWR\\TDOA} & \xmark  & 1 & \xmark   & \xmark  & 3D  & DWM1000      \\
   
   \citep{raza2019dataset}            & TDOA    & \xmark   & 1   & \xmark   & \xmark  & 2D  & DWM1001      \\
    \textbf{\makecell{UTIL Dataset\\(ours)}}  & TDOA    & \cmark   & 4     & \makecell{Plastic, wood,\\ metal, cardboard,\\ and foam} & Metal  & 3D & DWM1000  \\
  \bottomrule
  \end{tabularx}
  \label{tab:dataset_comp}
\end{table*}
Many public UWB datasets have been produced in literature for a variety of applications, including UWB radar~\citep{zhengliang2021dataset,ahmed2021uwb,ge2023large,zhang2023multi,brishtel2023driving}, human motion tracking~\citep{delamare2020new, vleugels2021ultra,bocus2022comprehensive}, localization for mobile robots~\citep{raza2019dataset, queralta2020uwb, nguyen2021ntu, moron2023benchmarking}, etc. In addition to these specific dataset papers, several UWB datasets have been publicly released as companions to research papers~\citep{bregar2018improving,barral2019nlos,ledergerber2019ultra,pfeiffer2021computationally}. Considering the large amount of applications of UWB technology, we focus on UWB-based localization for mobile robots and summarize the related UWB datasets, to the best of our knowledge, in Table~\ref{tab:dataset_comp} for an overview. 

From Table~\ref{tab:dataset_comp}, we can observe that many of the public UWB datasets focus on UWB TWR-based localization~\citep{bregar2018improving,li2018accurate,barral2019multi,arjmandi2020benchmark,nguyen2021ntu,moron2023benchmarking}. For UWB TDOA-based positioning, Raza \textit{et al.} \citep{raza2019dataset} provided a dataset to compare the performance of UWB TDOA and narrowband Bluetooth-based localization technologies. However, the dataset only contains UWB and Bluetooth radio measurements, lacking other sensing modalities. Additionally, the data collection occurred in a simple 2D setup without any obstacles. Pfeiffer \textit{et al.}  \citep{pfeiffer2021computationally} released their dataset along with their research work including both UWB TWR and TDOA measurements collected from a Crazyflie 2.1 nano-quadrotor. However, the dataset was also created in an obstacle-free environment.

% summarize the drawbacks for current datasets
Realistic indoor environments often contain different types of obstacles that might interfere with UWB radio signals. In order to achieve accurate and reliable UWB TDOA-based indoor positioning, UWB measurements need to be tested in such scenarios. Currently, there exists an absence of UWB TDOA measurement identification and data collection in a cluttered environment. Furthermore, most of these datasets do not provide data collected in the presence of dynamic obstacles. We present an Ultra-wideband Time-difference-of-arrival Indoor Localization (UTIL) dataset to fill in this gap. In this dataset, we conducted extensive UWB TDOA identification experiments under LOS and NLOS scenarios and collected multimodal sensor data from a quadrotor platform in the presence of static and dynamic obstacles. During the flight experiments, we collected raw UWB TDOA measurements with additional onboard sensor data (IMU, optical flow, and ToF laser) in four anchor constellations. Both centralized and decentralized TDOA measurements were collected under the same conditions for comparison. 
% A millimeter-accurate motion capture system is used to provide the ground truth data. 
%
The combination of multimodal onboard sensors, different anchor constellations, two TDOA modalities, and diverse cluttered scenarios contained in this dataset facilitates in-depth comparisons of UWB TDOA-based quadrotor localization capabilities, a level of analysis not achievable with existing datasets. To the best of our knowledge, a comprehensive UWB TDOA dataset with \textit{(i)} identification experiments and \textit{(ii)} data taken in a variety of indoor environments with static and dynamic obstacles does not exist in literature.

\section{UWB TDOA-based Localization System}
\label{sec:uwb_system}
Our UWB TDOA-based localization system is sketched in Figure~\ref{fig:uwb-system}. Eight UWB anchors were pre-installed in the space with known positions. The robot equipped with an UWB tag computes the difference of the distances between the robot and the two transmitting anchors using the UWB signal arrival times. To better explain the content of our dataset, we introduce the UWB radio hardware used for data collection and provide a brief explanation of the TDOA principles.

\subsection{UWB Sensor}
\begin{figure}[!tb]
  \begin{center}
    \includegraphics[width=.32\textwidth]{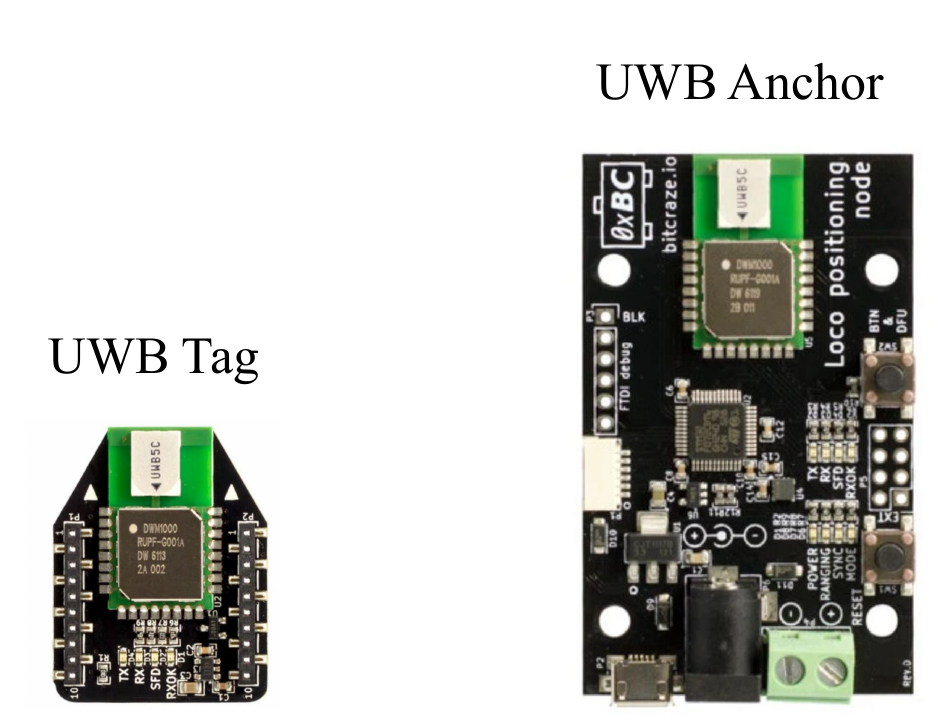}
  \end{center}
  \caption{The Loco Positioning System (LPS) anchor and tag from Bitcraze, based on Decawave's DWM1000 UWB modules, is used for the data collection.}
  \label{fig:dw1000}
\end{figure}

The Decawave's DWM1000~\citep{Datasheet} UWB radio sensors were used to create this dataset. The DWM1000 module is a low-cost UWB radio often used to develop cost-efficient localization solutions. The UWB TDOA measurements were collected using the Loco Positioning System (LPS) from Bitcraze, which is based on DWM1000 UWB modules (see Figure~\ref{fig:dw1000}).

Both centralized TDOA and decentralized TDOA~\citep{meng2016optimal} are implemented in LPS, which are referred to as TDOA 2 and TDOA 3 by Bitcraze, respectively. In centralized TDOA systems, all the anchors are synchronized w.r.t. one master anchor and the TDOA measurements are expressed in the same clock. However, centralized TDOA systems are limited by communication constraints and suffer from single point failure~\citep{ennasr2016distributed}. In decentralized TDOA systems, anchor pairs synchronize the timescales between each other and not with a single master anchor, which leads to scalability.

\subsection{Time-difference-of-arrival Principles}
\label{sec:tdoa_principle}
\begin{figure}[!tb]
    \begin{center}
	\includegraphics[width=.45\textwidth]{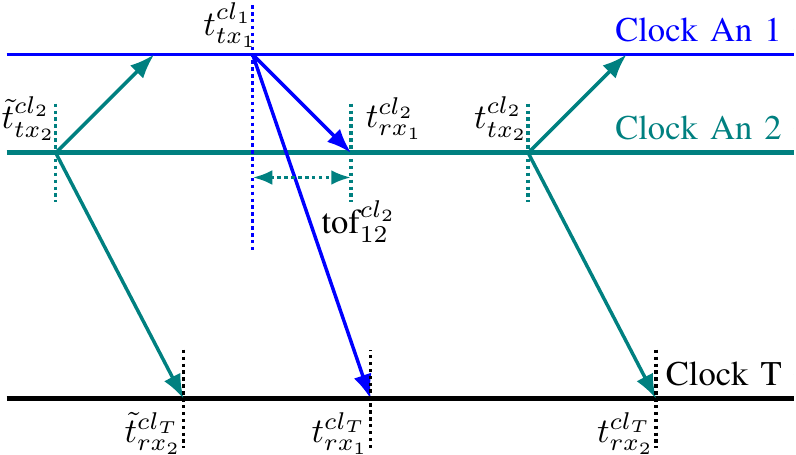}
    \end{center}
    \caption{The sequence of UWB radio packets between the tag and anchor 1 and anchor 2. The clocks of anchor 1, anchor 2, and the tag are indicated as solid lines with different colors. The radio packets between the tag and anchors are denoted as solid arrow lines. }
\label{fig:tdoa-principle}
\end{figure}

In this subsection, we briefly explain the TDOA principles implemented in LPS. Without loss of generality, we denote a pair of UWB anchors as anchor 1 and anchor 2. The TDOA measurement $d_{12}$ is the difference of distances of the tag to anchors 1 and 2. The sequence of the UWB radio packets among the anchor pair and one tag is visualized in Figure~\ref{fig:tdoa-principle}. UWB radios, including both anchors and the tag, operate independently with their own individual clocks. These clocks are depicted as solid lines, each distinguished by a unique color. A clock synchronization process is essential for an accurate computation of TDOA measurements. 

% TDOA principle figure
We denote the clock of anchor 1, anchor 2, and the tag as $cl_1$, $cl_2$, and $cl_T$ in the superscripts. The transmission and the reception of a radio signal from anchor 1 are denoted as $tx_1$ and $rx_1$ in the subscripts. As an example, $t_{rx_2}^{cl_T}$ indicates the receiving timestamp of the radio packet from anchor 2 expressed in the tag's clock. UWB radio signals are transmitted with a scheduled transmission sequence. We use $\tilde{t}$ to indicate the timestamp from the previous sequence: $\tilde{t}_{tx_2}^{cl_2}$ indicates the transmitting timestamp of the radio packet from anchor 2 expressed in the clock of anchor 2 from the previous sequence. With anchor 1 and anchor 2 at positions $\bm{a}_1,\bm{a}_2\in\mathbb{R}^3$ and one tag at position $\mathbf{p}\in\mathbb{R}^3$, the TDOA measurement between anchor 1 and 2 is computed as 
\begin{equation}
\begin{split}
    d_{12} &= c\left[(t_{rx_2}^{cl_T} - t_{rx_1}^{cl_T}) - \alpha(t_{tx_2}^{cl_2} - t_{rx_1}^{cl_2} + \textrm{tof}^{cl_2}_{12})  \right]  \\
    & = \|\mathbf{p} - \bm{a}_2\| - \|\mathbf{p} - \bm{a}_1\|
\end{split},
    \label{eq:tdoa-equation}
\end{equation}
where $c$ indicates the speed of light, $\alpha$ is the clock correction parameter converting from anchor 2's clock to the tag's clock, $\textrm{tof}^{cl_2}_{12}$ is the time-of-flight measurements between anchor 1 and anchor 2 expressed in anchor 2's clock, and $\|\cdot\|$ indicates the $\ell_2$ norm. The clock correction parameter $\alpha$ is used to synchronize the clock of anchor 2 to the clock of the tag. We compute $\alpha$ with the timestamps from the previous sequence
\begin{equation}
    \alpha = \frac{t_{rx_2}^{cl_T} - \tilde{t}_{rx_2}^{cl_T}}{t_{tx_2}^{cl_2} - \tilde{t}_{tx_2}^{cl_2}}.
    \label{eq:clock correction}
\end{equation}

\section{Data Collection}
\label{sec:data_collection}
\begin{figure}[!t]
    \centering
    \includegraphics[width=.35\textwidth]{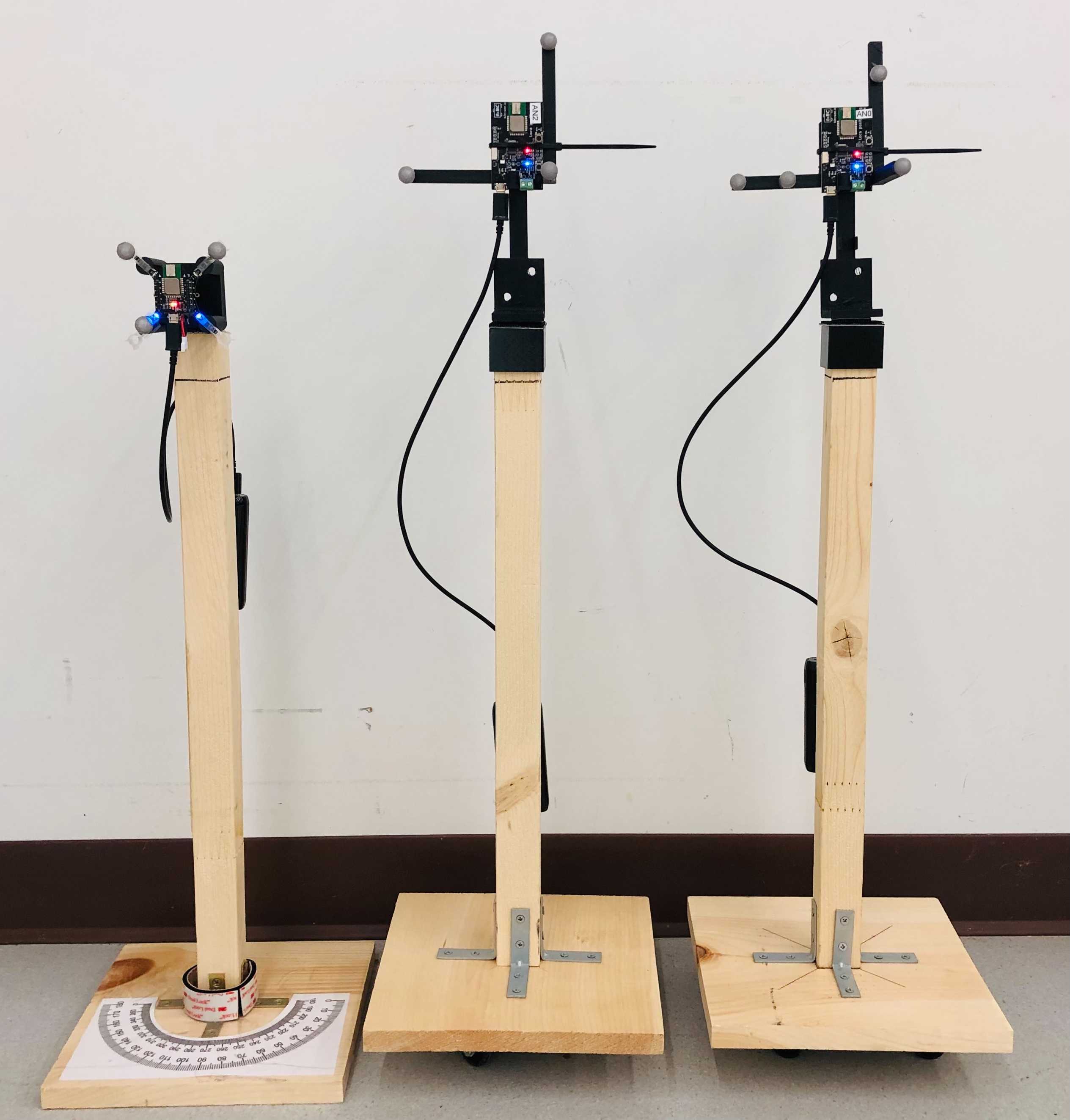}
    \caption{UWB anchors and tag setup for UWB identification experiments.}
    \label{fig:static_setup}
\end{figure}

\subsection{Ground Truth}
The UTIL dataset was created at the University of Toronto Institute for Aerospace Studies (UTIAS). We collected the data in the indoor flight arena equipped with a motion capture system of $10$ Vicon Vantage+ cameras~\citep{viconCam}. The millimeter-level accurate Vicon pose measurements were collected during UWB identification experiments and flight experiments as the ground truth measurements.

\begin{figure}[!b]
  \begin{center}
    \includegraphics[width=.48\textwidth]{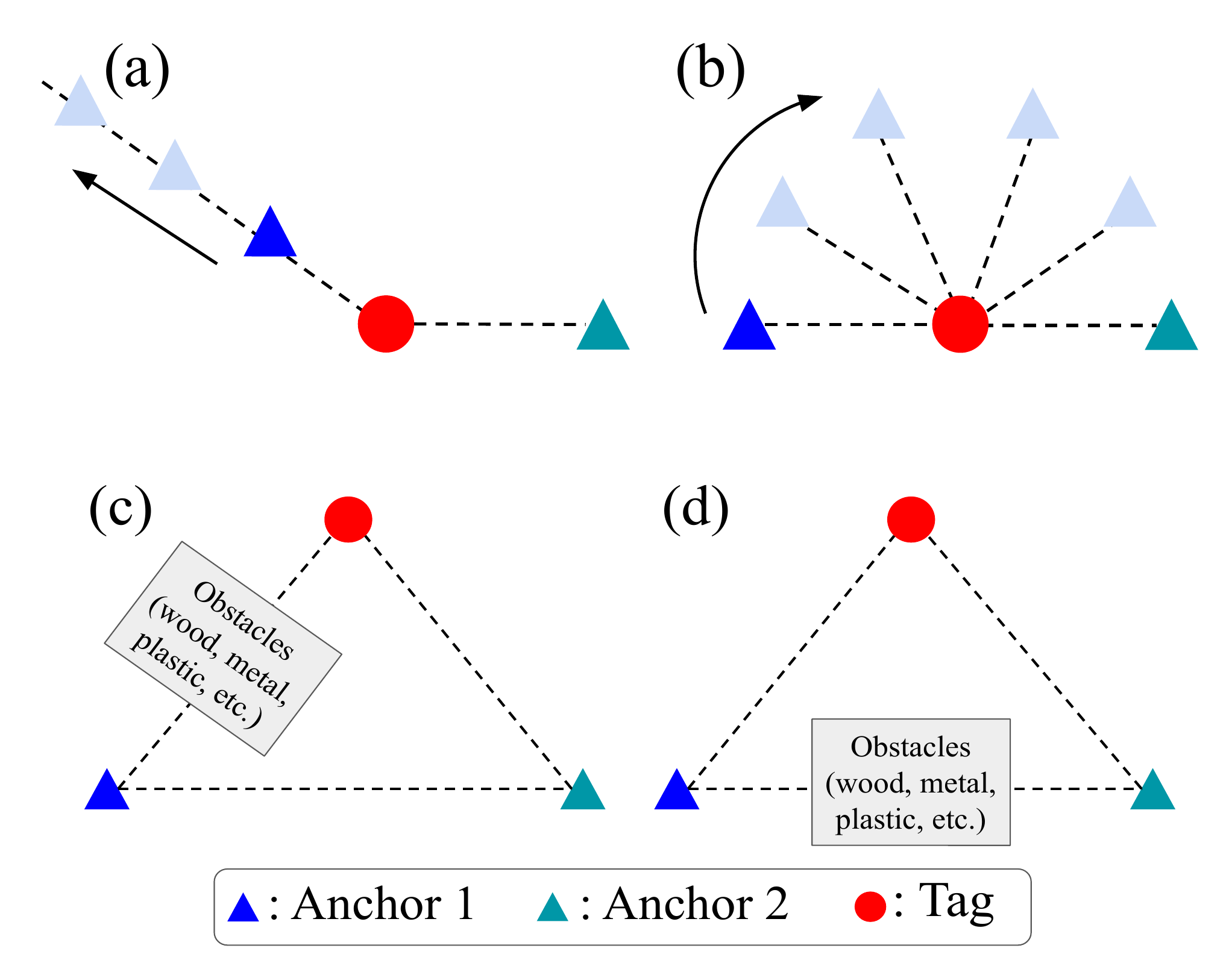}
  \end{center}
  \caption{A diagram of the UWB identification experiments. The experiment process of LOS distance tests and LOS angle tests are illustrated in (a) and (b). The NLOS identification tests between an anchor and a tag and between two anchors are shown in (c) and (d). }
  \label{fig:identify_exp}
\end{figure}
\subsection{UWB Identification Dataset}
In order to identify the UWB TDOA measurement performance in LOS and NLOS scenarios, we conducted a variety of LOS and NLOS experiments using two anchors and one tag. Figure~\ref{fig:static_setup} demonstrates the experimental setup for the identification dataset. Two UWB anchors referred to as anchor 1 and anchor 2, and one Crazyflie nano-quadrotor equipped with an UWB tag were placed on wooden structures. The ground truth pose data was provided by the motion capture system. Since only two anchors were used for the data collection, we ignored the difference between the centralized and decentralized TDOA modes and set the anchors into decentralized mode (TDOA 3). We collected the data through the Robot Operating System (ROS). Each sub-dataset was collected during a one-minute static experiment. 

To assess the quality of received UWB signals, we collected the signal-to-noise ratio (SNR) and power difference ($P_d$) values provided in the Decawave user manual~\citep{UserManual} as the performance metrics. The computation of SNR and power difference $P_d$ are as follows
\begin{equation}
    \textrm{SNR} = \frac{\textrm{Am}_f}{\sigma_n}, ~~~~~ P_d  = P_r - P_f,
\end{equation}
where $\textrm{Am}_f$ indicates the \textit{First Path Amplitude value}, $\sigma_n$ indicates the \textit{Standard Deviation of Channel Impulse Response Estimate Noise} value, and $P_r$ and $P_f$ are the total received power and the first path power, respectively. In general, a higher SNR value or a lower $P_d$ indicates the received radio signal is of good quality~\citep{UserManual}. The four raw measurements $\{\textrm{Am}_f, \sigma_n,  P_r, P_f\}$ can be accessed from the DW1000 UWB chip when receiving an UWB radio signal. We refer to Section 4.7 of the user manual~\citep{UserManual} for more information. Detailed data format and descriptions of the UWB identification dataset are provided in Section~\ref{sec:identify_data_format}. 
%  LOS box plots
\begin{figure}[!t]
    \centering
    \includegraphics[width=0.48\textwidth]{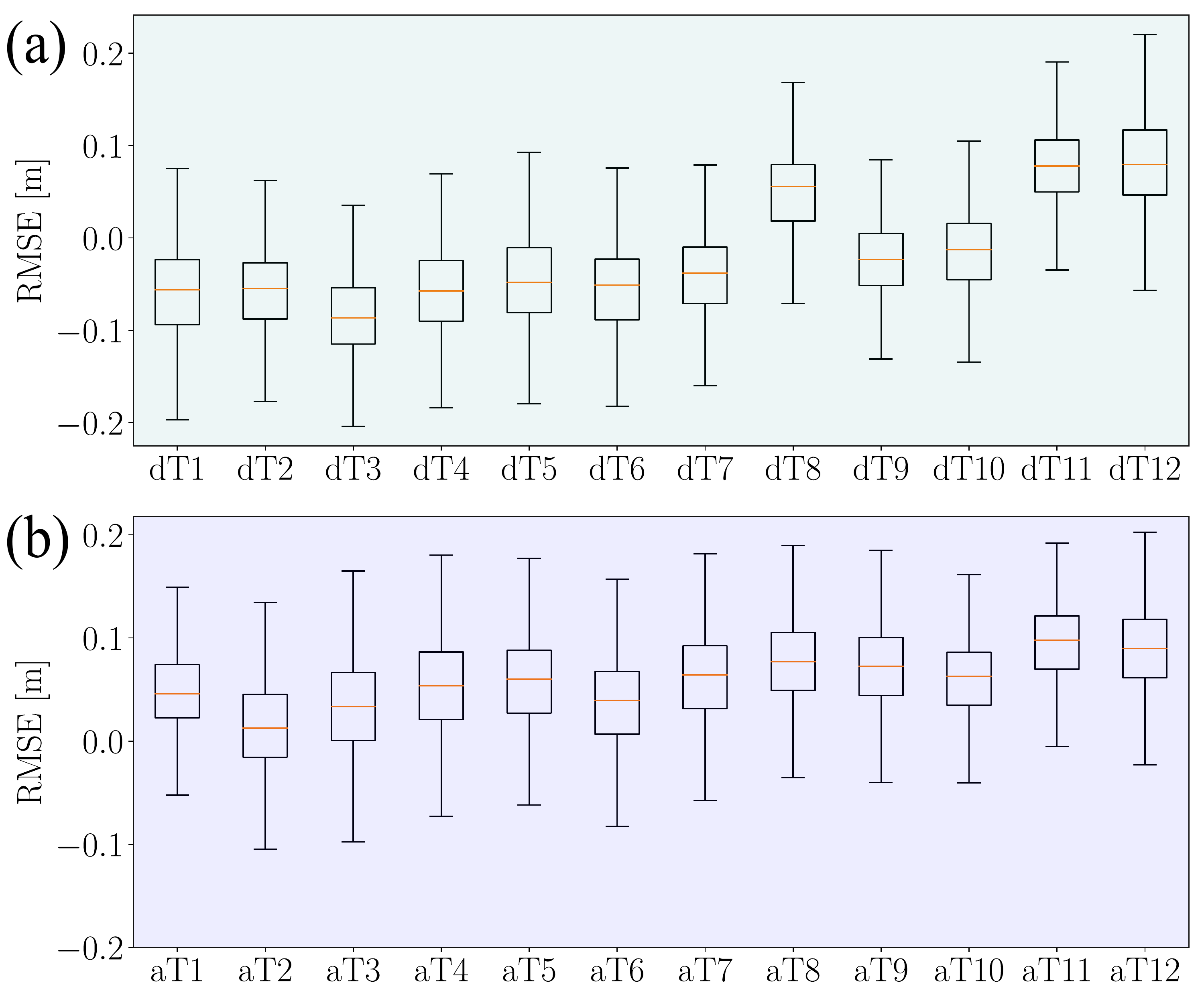}
    \caption{Measurement errors in (a) LOS distance tests (top) and (b) LOS angle tests (bottom). We indicated the distance test and angle tests as dT$\#$ and aT$\#$, where $\#$ is the test number.}
    \label{fig:los_err}
\end{figure}

%  NLOS box plots
\begin{figure}[!b]
    \centering
    \includegraphics[width=.48\textwidth]{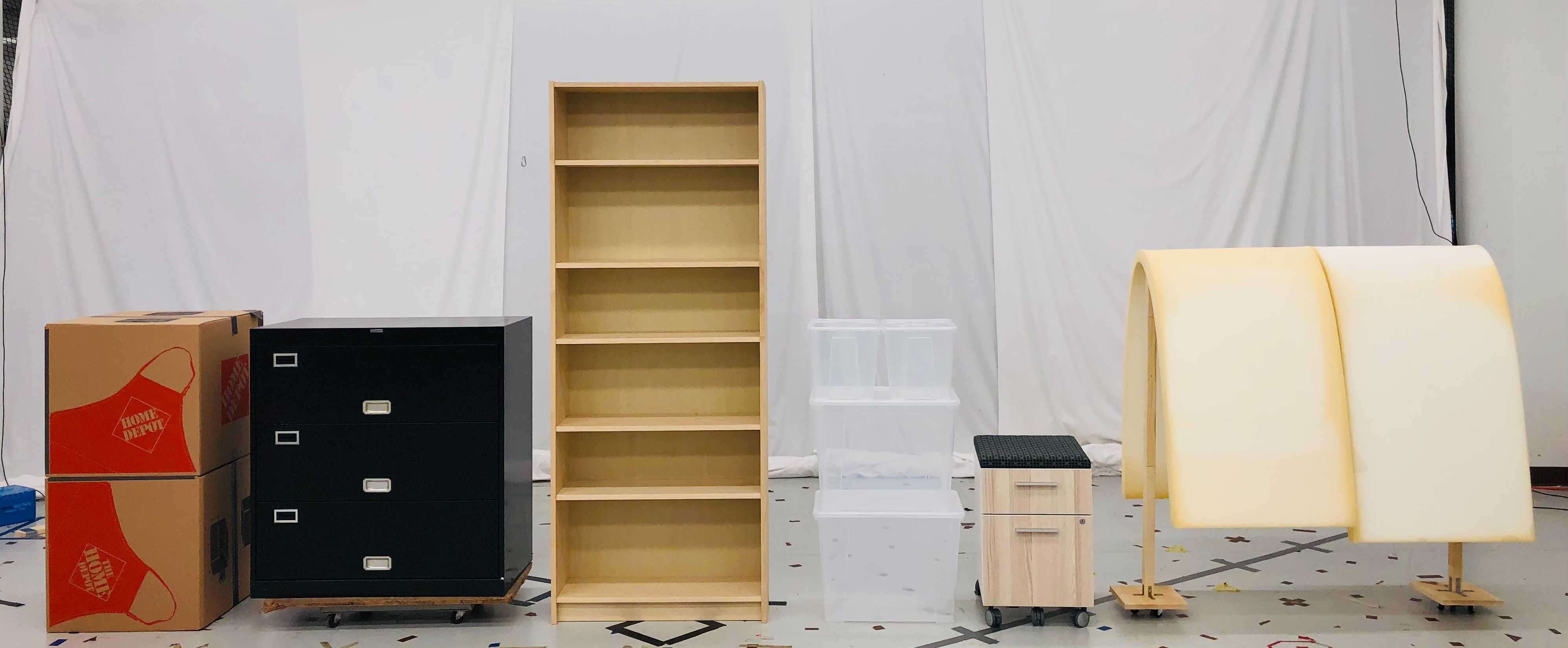}
    \caption{Obstacles used in the NLOS tests. The materials of the obstacle from left to right are cardboard, metal, wood, plastic, wood, and foam. }
    \label{fig:obstacles}
\end{figure}
\subsubsection{Line-of-sight Tests.} We collected UWB TDOA measurements from two LOS identification tests: \textit{(i)} the LOS distance test and \textit{(ii)} the LOS angle test. The data collection procedures are sketched in Figure~\ref{fig:identify_exp}a-b. The positions of the tag and anchor $2$ were fixed throughout the LOS data collection process. In LOS distance test, we increased the distance between anchor $1$ and the tag from $0.5$ meters to $6.5$ meters in intervals of $0.5$ meters. In LOS angle test, we changed the angle between two anchors from $180^{\circ}$ to $15^{\circ}$ in intervals of $15^{\circ}$. The LOS TDOA measurement errors are summarized in Figure~\ref{fig:los_err}. We indicate the distance test and angle tests as dT$\#$ and aT$\#$, where $\#$ is the test number. 

\begin{figure}[!t]
    \centering
    \includegraphics[width=0.5\textwidth]{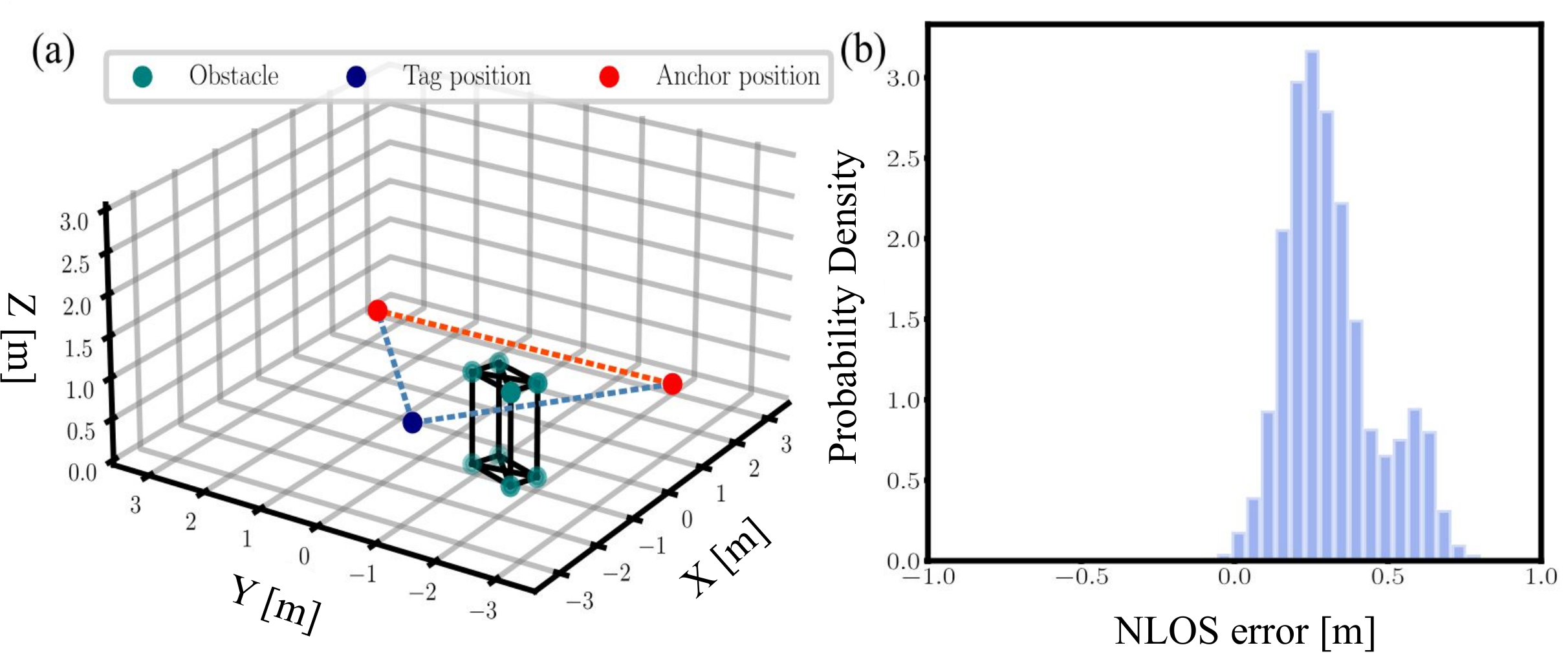}
    \caption{One example of NLOS identification experiments is shown in (a). A histogram of measurement errors induced by placing the metal obstacle between one anchor and the tag is summarized in (b). }
    \label{fig:nlos_err}
\end{figure}
\subsubsection{Non-line-of-sight Tests.}
In the NLOS identification tests, we fixed the positions of the tag and two anchors and placed different types of obstacles to create NLOS scenarios. Four reflective markers were placed on the top surface of the obstacle to capture both its position and dimensions during the experiments. To reflect the comprehensive performance of UWB NLOS measurements, we selected six obstacles of different types of materials commonly used in indoor settings, including cardboard, metal, wood, plastic, and foam. Figure~\ref{fig:obstacles} shows the obstacles we used during the experiments. 

% flying arena
\begin{figure}[!b]
    \centering
    \includegraphics[width=0.48\textwidth]{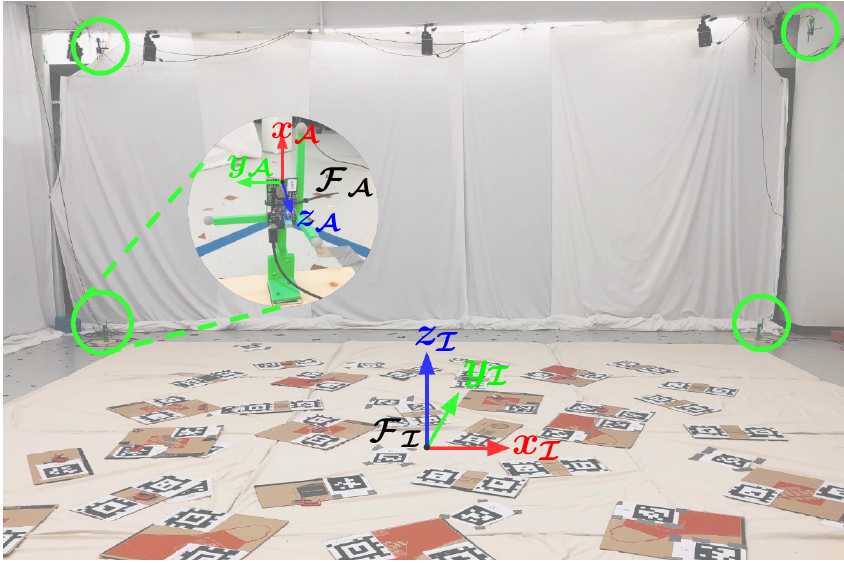}
    \caption{A photo of the flight arena. The UWB anchors are enclosed with green circles. The inertial frame and UWB anchor frame are indicated as $\mathcal{F}_\mathcal{I}$ and $\mathcal{F}_\mathcal{A}$, respectively. The ground is covered with soft mattresses with a thickness of two inches ($5.08$ cm). Printed AprilTags are attached to the mattresses to provide visual features for the optical flow.}
    \label{fig:fight_arena}
\end{figure}
As explained in Section~\ref{sec:tdoa_principle}, the UWB tag listens to the radio packets transmitting between anchors to compute TDOA measurements. Both NLOS conditions between one anchor and the tag and between two anchors will affect TDOA measurements. Therefore, we conducted NLOS experiments under \textit{(i)} NLOS conditions between anchor $1$ and the tag and \textit{(ii)} NLOS conditions between anchor $1$ and anchor $2$ (see Figure~\ref{fig:identify_exp}c-d). Considering the different radio reflection and diffraction effects with one obstacle under different orientations, we collected six sub-datasets for each NLOS condition with different orientations of the obstacle. One LOS dataset was collected for comparison. We present one NLOS identification experiment and summarize the measurement errors induced by metal occlusions in Figure~\ref{fig:nlos_err} as an example.

\subsection{Flight Dataset}
The flight dataset is a comprehensive multimodal collection from a customized quadrotor platform in a range of cluttered indoor environments, featuring both static and dynamic obstacles. 

\subsubsection{Indoor Flight Arena.} We collected the UWB TDOA flight dataset in the indoor flight arena measuring approximately $7.0~m\times8.0~m\times3.5~m$. Printed AprilTags~\citep{olson2011apriltag} were attached to the soft mattresses to provide visual features for optical flow. Figure~\ref{fig:fight_arena} is a photograph of our flight arena during data collection. 

% % flight traj.
\begin{figure*}[!t]
    \centering
    \includegraphics[width=0.98\textwidth]{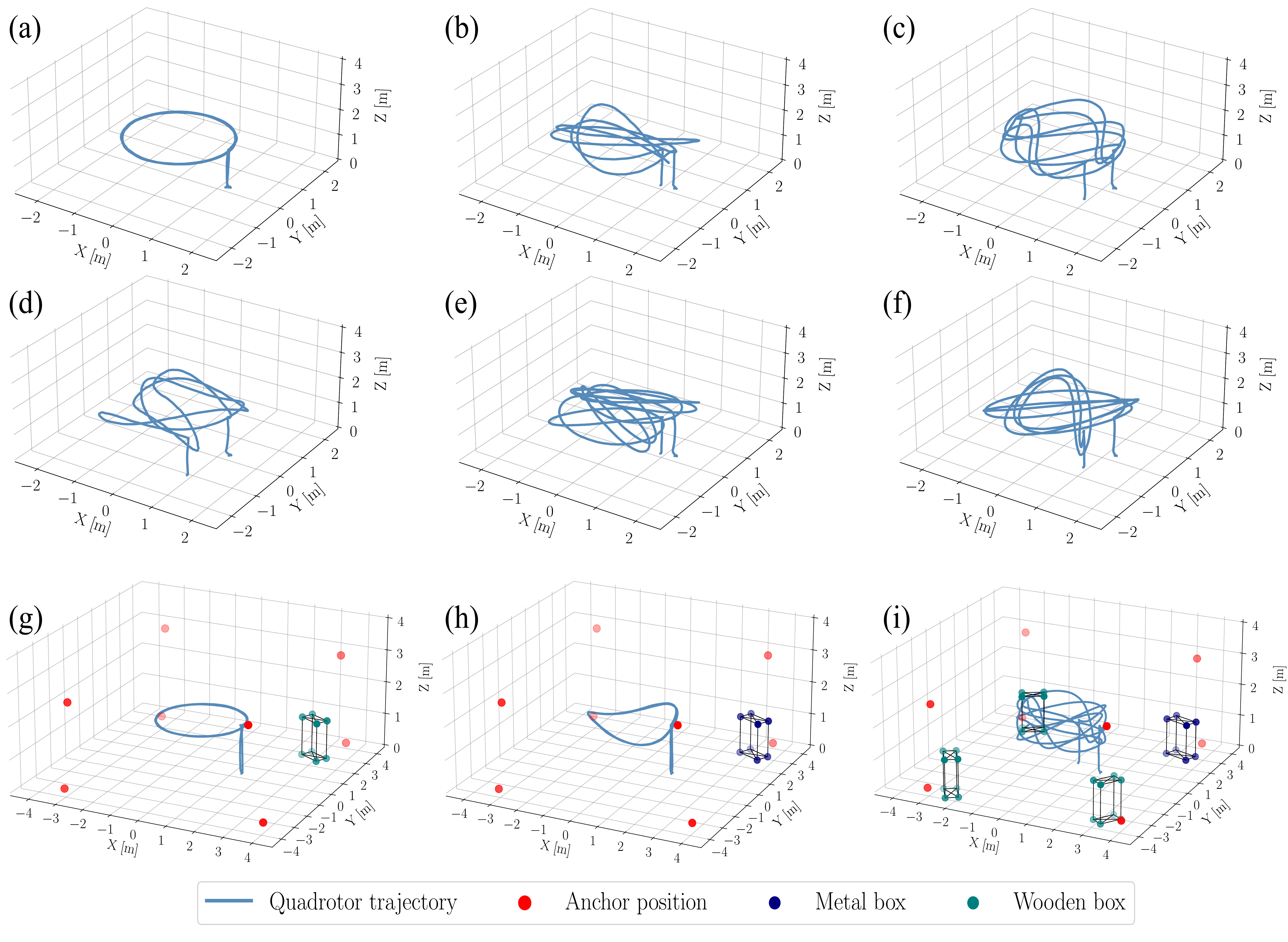}
    \caption{Flight trajectories and static NLOS conditions in the flight dataset. The six flight trajectories in constellation $\#1$, $\#2$, and $\#3$ are shown in (a)-(f). Note that the trajectories in constellation $\#1$ have a smaller separation in the x and y axes due to the smaller constellation coverage. The three static NLOS conditions and the anchor positions in constellation $\#4$ together with the three flight trajectories are shown in (g)-(i).}
    \label{fig:flight-exp-traj}
\end{figure*}

To ensure the accuracy of our dataset, we performed anchor surveying for each of our data collection sessions, leading to four different anchor constellations. In each anchor constellation, eight UWB anchors were pre-installed in the flight arena.  We refer to the Vicon frame as the inertial frame $\mathcal{F}_\mathcal{I}$ and the anchor frame as $\mathcal{F}_\mathcal{A}$ (see Figure~\ref{fig:fight_arena}). To maximize the utilization of the indoor environment, we placed the UWB anchors along the boundary of the space to construct the anchor constellations, resulting in their positions outside the field of view of the Vicon system. Consequently, we used a millimeter-accurate Leica total station~\citep{Leica} for the anchor survey process and transformed the survey results back into the Vicon inertial frame subsequently for use. To align the total station frame and the inertial frame, we used the total station to survey six reflective markers with known positions in the inertial frame and compute the transformation matrix by aligning the point-clouds~\citep{besl1992method}. To assess the quality of the frame alignment between the total station frame and the inertial frame, we computed the reprojection error of the six reflective markers. The survey points in the total station frame were converted into the inertial frame with a root-mean-squared error (RMSE) of around $1.12$ mm. Since the low-cost DWM1000 UWB chip is reported to have pose-related measurement biases~\citep{zhao2021learning,ledergerber2017ultra}, we intended to survey both the position and the orientation of each anchor for reproducibility. During the anchor surveying process, we surveyed the UWB antenna center together with three reflective markers on the extended arms of each anchor stand (see Figure~\ref{fig:fight_arena}). The positions of these markers in the anchor frame were pre-measured. Then we leveraged the surveyed marker positions, converted into the inertial frame, along with their known positions in the anchor frame to compute the pose (position and orientation) of each anchor through point-cloud alignment~\citep{besl1992method}. We provide both Python (\texttt{anchor\_survey.py}) and MATLAB (\texttt{anchor\_survey.m}) scripts in our development kit for the users to replicate this process.

\subsubsection{Quadrotor Platform.}
\begin{figure}[!t]
    \centering
    \includegraphics[width=0.48\textwidth]{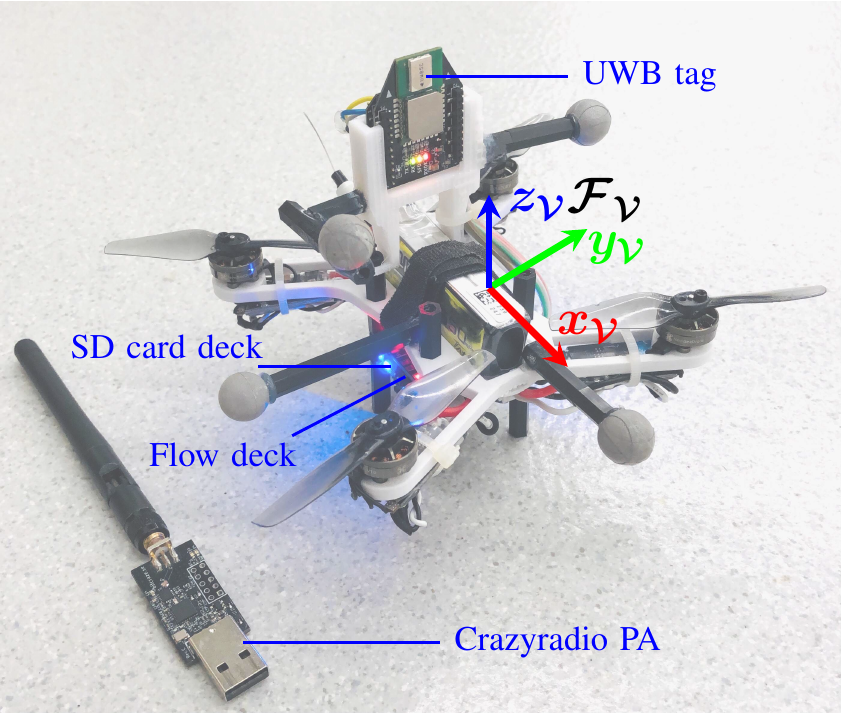}
    \caption{The customized quadrotor platform based on the Crazyflie Bolt flight controller.}
    \label{fig:customize-drone}
\end{figure}

We built a customized quadrotor based on the Crazyflie Bolt flight controller~\citep{cfbolt} with an inertial measurement unit (IMU) and attached commercially available extension boards (so-called decks) from Bitcraze for data collection (see Figure~\ref{fig:customize-drone}). The LPS UWB tag was mounted vertically on the top since the DWM1000 antenna radiation pattern is uniform in its azimuth plane~\citep{Datasheet}.  A flow deck attached at the bottom provides optical flow measurements and a laser-based time-of-flight (ToF) sensor provides the local altitude information. The accelerometer and gyroscope data were obtained from the onboard IMU. A micro SD card deck was used to log the raw sensor data received by the flight control board with high-precision microsecond timestamps. The customized quadrotor communicates with a ground station computer over a 2.4 GHz USB radio dongle (Crazyradio PA) for high-level interaction. In terms of software, we used the Crazyswarm package~\citep{preiss2017crazyswarm} to send high-level commands, such as takeoff/landing and start/stop of data logging, and pre-defined waypoints. The pose of the quadrotor measured by the motion capture system was also sent to the quadrotor as an external measurement for the onboard state estimation. 
% quadrotor

\subsubsection{Calibration and Latency.}
% calibration
We refer to the offset between the center of a sensor and the center of vehicle frame $\mathcal{F}_\mathcal{V}$ as sensor extrinsic parameters. We calibrated the sensor extrinsic parameters by manually measuring the translation vectors from the vehicle center to onboard sensors (UWB tag and flow deck). The IMU was assumed to be aligned with the vehicle center. The translation vector from the vehicle to the UWB tag was measured as $r_{uv} = [-0.012, 0.001, 0.091]^T$~m and the measurement model is: 
\begin{equation}
    d_{ij} = \|\left(\mathbf{C}_{\mathcal{IV}}r_{uv} + \mathbf{p}\right) - \bm{a}_j\| -  \|\left(\mathbf{C}_\mathcal{IV}r_{uv} + \mathbf{p}\right) - \bm{a}_i\|,
\end{equation}
where $\mathbf{C}_\mathcal{IV}$ is the rotation matrix from the vehicle frame $\mathcal{F}_\mathcal{V}$ to the inertial frame $\mathcal{F}_\mathcal{I}$ and $\mathbf{p}$ indicates the position of the vehicle expressed in the inertial frame. 

Similarly, the translation vector from the vehicle to the flow-deck extension board was measured to be $r_{fv} = [0.000, 0.000, -0.012]^T$~m. Since we covered the ground of the flight arena with 2-inch thick ($0.0508$~m) mattresses for protection during data collection, the thickness of the mattress needs to be taken into account while using the ToF measurements. We refer to Section 6.5 of \citep{greiff2017modelling} for detailed information on the flow-deck extension board.

% latency
Given our adoption of the same software and hardware configuration as the Crazyswarm project~\citep{preiss2017crazyswarm} for our flight experiments, we direct readers to Section XI-A of~\citep{preiss2017crazyswarm} for a comprehensive explanation of the latency measurement process, in which the reported latency for a single vehicle is approximately $11$ ms. 

%%% table
\begin{table}[!b]
\centering
\caption{Format of the \texttt{.csv} files in each static sub-dataset.}
\label{table:static_data_content}
\renewcommand{\arraystretch}{1.7}
\small
\begin{tabular}{ | c | c | l | } 
\hline
 \makecell[c]{CSV \\Column} & Value       & \makecell[c]{Description}    \\ 
\hline
1      & $d_{12}$ [m] &TDOA measurements $d_{12} = d_2 - d_1$         \\ 
\hline
2      & $d_{21}$ [m] &TDOA measurements $d_{21} = d_1 - d_2$        \\ 
\hline 
3      & $\textrm{SNR}_1$   & \makecell[l]{SNR value of the UWB radio\\ packet sent from anchor $1$ \\received by the tag}  \\ 
\hline
4      & $P_{d1}$ [dB] & \makecell[l]{power difference value of the UWB \\radio packet sent from anchor $1$ \\ received by the tag} \\
\hline
5      & $\textrm{SNR}_2$   & \makecell[l]{SNR value of the UWB radio\\ packet sent from anchor $2$ \\received by the tag}  \\
\hline
6      & $P_{d2}$ [dB] & \makecell[l]{power difference value of the UWB \\ radio packet sent from anchor $2$ \\ received by the tag} \\
\hline
7      & $\textrm{SNR}^{an1}$  & \makecell[l]{SNR value of the UWB radio\\ packet sent from anchor $2$ \\received by anchor $1$} \\
\hline
8      & $P_{d}^{an1}$ [dB]& \makecell[l]{power difference value of the UWB \\ radio packet sent from anchor $2$ \\ received by anchor $1$ } \\
\hline
9      & $r_{12}^{cl1}$ [m] & \makecell[l]{distance between anchor $1$ and $2$ \\computed by $\textrm{tof}_{12}^{cl1}$ }\\
\hline
10     & $\textrm{SNR}^{an2}$  & \makecell[l]{SNR value of the UWB radio\\ packet sent from anchor $1$ \\received by anchor $2$ } \\
\hline
11     & $P_{d}^{an2}$ [dB] & \makecell[l]{power difference value of the UWB \\ radio packet sent from anchor $1$ \\ received by anchor $2$} \\
\hline
12     & $r_{12}^{cl2}$ [m]& \makecell[l]{distance between anchor $1$ and $2$ \\computed by $\textrm{tof}_{12}^{cl2}$ }\\
\hline
\end{tabular}
\end{table}

\subsubsection{Data Collection Process.}
We operated the motion capture system at a fixed sample frequency of $200$ Hz and sent the measured quadrotor pose to the onboard error-state Kalman filter with a small standard deviation, $0.001$ m for position and $0.05$ rad for orientation, for state estimation. Onboard the quadrotor, the raw UWB TDOA measurements, gyroscope, accelerometer, optical flow, ToF laser-ranging, barometer, and the Vicon pose measurements (sent from the ground station) were recorded as event streams. We treat the Vicon pose measurements logged onboard as the ground truth data. Each datapoint was timestamped with the onboard microsecond timer and the resulting time series were written to the micro SD card as a binary file. Python scripts are provided to parse and analyze the binary data.

During the data collection process, we commanded the quadrotor to fly six different trajectories in constellation $\#1$, $\#2$, and $\#3$ under LOS conditions. The six flight trajectories are summarized in Figure~\ref{fig:flight-exp-traj}a-f. In constellation $\#4$, we created three cluttered environments with static obstacles (see Figure~\ref{fig:flight-exp-traj}g-i) and two cluttered environments with one dynamic metal obstacle. During the dynamic obstacle experiment, we moved the metal cabinet manually and intentionally blocked two anchors temporarily during the flights. Note that the human body also acted as a dynamic obstacle in these cases. The onboard sensor data was collected over three different trajectories. In each experiment, we commanded the quadrotor to fly the same trajectories with both centralized and decentralized TDOA modes for comparison. For the flight experiments with dynamic obstacles, we created animations in \texttt{scripts/flight-dataset/animations} folder to demonstrate the data collection process. We also conducted a couple of manual data collection experiments with a human body involved in constellation $\#3$ and $\#4$.

\begin{table*}[!t]
\centering
\caption{Summary of the CSV flight dataset format. }
\label{table:flight_dataset}
\renewcommand{\arraystretch}{1.2}
\begin{tabular}[0.8*\textwidth]{ |c |l | l |} 
\hline
CSV column &Name           & Format     \\ 
\hline
$1\sim4$ & UWB TDOA       & (timestamp [ms], Anchor-ID $i$,  Anchor-ID $j$, $d_{ij}$ [m])     \\ 
\hline
$5\sim8$  & Acceleration   & (timestamp [ms], acc. $x$ [G],  acc. $y$ [G], acc. $z$ [G])      \\ 
\hline
$9\sim12$  & Gyroscope      & (timestamp [ms], gyro. $x$ [deg/s],  gyro. $y$[deg/s], gyro. $z$[deg/s])      \\ 
\hline
$13\sim14$  & ToF laser-ranging    & (timestamp [ms], ToF [m])      \\ 
\hline
$15\sim17$  & Optical flow   & (timestamp [ms], dpixel $x$,  dpixel $y$)      \\ 
\hline
$18\sim19$ & Barometer      & (timestamp [ms], barometer [asl])      \\ 
\hline
$20\sim27$ & Ground truth pose  & \makecell[l]{(timestamp [ms], $x$ [m], $y$[m], $z$[m], $q_x$, $q_y$, $q_z$, $q_w$)}     \\ 
\hline
\end{tabular}
\end{table*}

\section{Data Format}
\subsection{UWB Identification Data Format}
\label{sec:identify_data_format}
The UWB identification dataset consists of data collected from \textit{(i)} LOS distance and angle tests and \textit{(ii)} NLOS identification experiments with different types of obstacles. In each sub-dataset, we provide a \texttt{.csv} file containing the collected data and a \texttt{.txt} file containing the poses of the tag and two anchors in one folder. For NLOS identification experiments, the positions of the four markers on the obstacles are also included in the \texttt{.txt} file. The format of the \texttt{.csv} file and brief descriptions of each value are summarized in Table~\ref{table:static_data_content}. The format of the position data provided in the \texttt{.txt} file is $(x,y,z)$ in meters. The orientation is provided as a unit quaternion $(q_x, q_y, q_z, q_w)$, where $q_w$ and $(q_x, q_y, q_z)$ are the scalar and the vector components, respectively. 
\begin{figure}[!b]
    \centering
    \includegraphics[width=.45\textwidth]{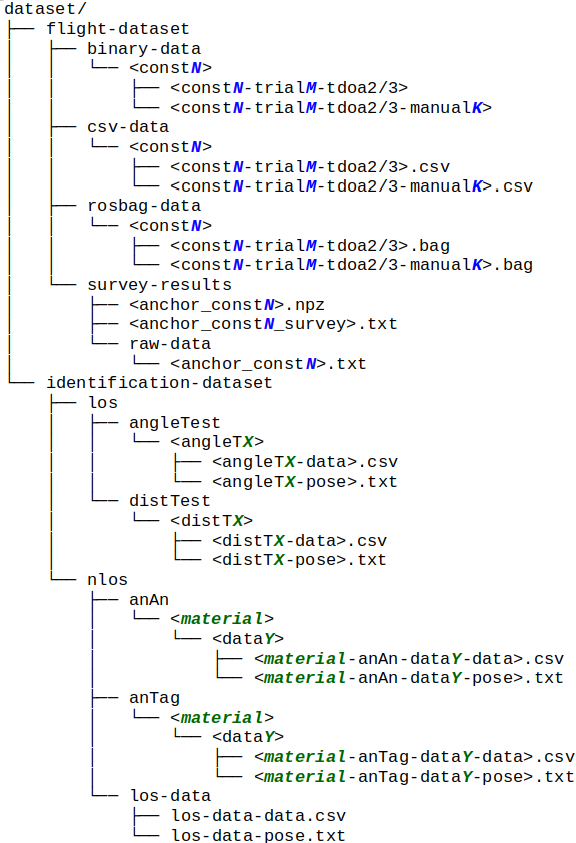}
    \caption{The file structure and naming convention of the UTIL dataset. In the flight dataset, we summarized the binary, CSV, and rosbag data according to different anchor constellations. In the identification dataset, we separate the LOS and NLOS testing data with each NLOS dataset containing the material of the obstacle in the filename.  }
    \label{fig:data-format}
\end{figure}

\subsection{Flight Experiments Data Format}
\label{sec:flight_data_format}
The flight experiment data were collected onboard the quadrotor during the flights as binary files. We provide the converted CSV and rosbag data and the corresponding Python scripts used for data parsing. For each UWB constellation, we provide the raw Leica total station survey results and computed anchor poses in \texttt{.txt} files. In each sub-dataset, we provide the timestamped accelerometer, gyroscope, UWB TDOA, optical flow, ToF laser-ranger, barometer measurements, and the ground truth measurements of the quadrotor's pose during the flight. The CSV data format for each sensor data is summarized in Table~\ref{table:flight_dataset}. The detailed file structure and naming convention are shown in Figure~\ref{fig:data-format}.

\section{Development Kit and Benchmark}
\label{sec:dev_kit}
As part of this dataset, we provide a development kit with both Python and MATLAB scripts for the users to parse the data. For the UWB identification dataset, we provide scripts to visualize the distribution of the collected data. For the flight dataset, we provide a range of different ways to visualize the sensor data and the data collection process. We also provide the STL files for the 3D printed quadrotor frame and the UWB tag support in the \texttt{setup\_files/stl-files} folder. Finally, an error-state Kalman filter (ESKF) implementation is provided for users to evaluate the UWB TDOA-based positioning performance in different scenarios. The development kit and instructions can be found at \url{https://utiasdsl.github.io/util-uwb-dataset/}. 

We provide a localization performance benchmark of the proposed dataset using IMU and UWB TDOA measurements. This evaluation is conducted with an error-state Kalman filter (ESKF)~\citep{goudar2021online} and a batch estimation algorithm~\citep{barfoot2017state}. A Chi-square test outlier rejection mechanism is applied in the ESKF to discard large outliers. The root-mean-square error (RMSE) in meters is summarized in Table~\ref{tab:benchmark_rmse}. The benchmark results reveal that ESKF and batch estimation demonstrates commendable performance with an approximate positioning error of $10$ cm in obstacle-free environments (constellations $\#1$, $\#2$, and $\#3$). However, in constellation $\#4$, the positioning performance of conventional estimation algorithms deteriorates greatly, primarily due to obstacle-induced measurement errors. These findings highlight substantial opportunities for researchers to develop novel estimation algorithms and enhance localization performance in cluttered indoor environments.

\section{Dataset Usage}
\label{sec:data_use}
In this section, we provide two potential usages of this dataset for users followed by a discussion of potential research directions. 

\subsection{UWB TDOA measurement modeling}
For UWB TDOA localization, modeling the measurement errors under LOS and NLOS scenarios is important for the design of localization algorithms~\citep{ruiz2017comparing,prorok2012online}. The stationary UWB TDOA signal testing data can be used to model the distribution of the UWB TDOA measurement errors under various LOS and NLOS conditions. One example of using the identification dataset to model UWB TDOA measurement error modeling can be found in~\cite{zhao2022finding}.

\subsection{Accurate UWB TDOA-based localization}
% % tdoa measurements
\begin{figure*}[!t]
    \centering
    \includegraphics[width=1.0\textwidth]{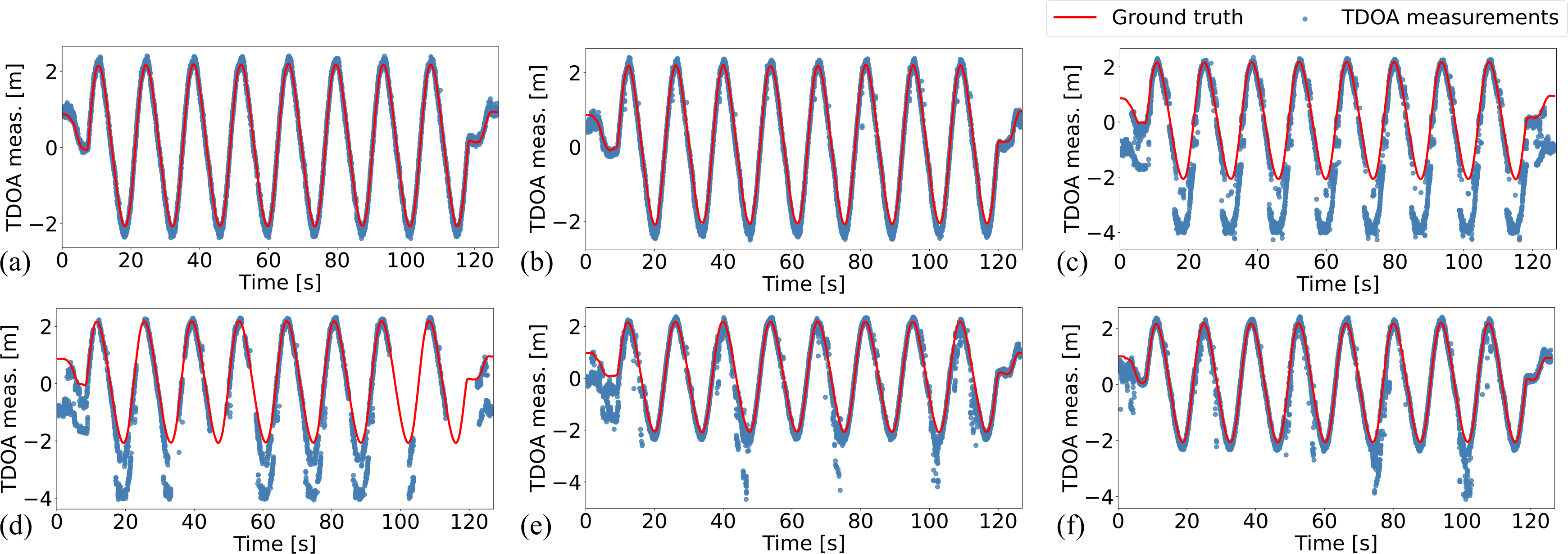}
    \caption{The UWB TDOA measurement $d_{23}$ collected from the same circle trajectory in constellation $\#4$ under different LOS/NLOS conditions. The measurements under clear LOS conditions are shown in (a). In static NLOS conditions induced by (b) one wooden obstacle, (c) one metal obstacle, and (d) one metal and three wooden obstacles, we can observe consistent measurement biases over repeated trajectories. In dynamic NLOS conditions caused by (e) one metal obstacle and (f) one metal obstacle and three wooden obstacles, the induced measurement errors are less predictable. }
    \label{fig:flight-meas-comp}
\end{figure*}
The flight dataset can be used to develop UWB TDOA-based localization algorithms. We provide the UWB measurements under centralized TDOA mode (TDOA2) and decentralized TDOA mode (TDOA3). The flight dataset collected in constellations 1 and 2 can be used to compare the localization performance between different UWB modes using low-cost DWM1000 UWB modules. 

%%% benchmark table
\begin{table}[!t]
\footnotesize
% \small
  \centering
  \setlength{\tabcolsep}{13.5pt}
  \renewcommand{\arraystretch}{1.0}
  \captionsetup{width=1.0\linewidth}
  \caption{Root-mean-square errors (RMSEs) in meters of obtained from benchmark on all sequences of the UITL dataset using IMU and UWB TDOA measurements. This evaluation is conducted using an error-state Kalman filter (ESKF) and a batch estimation algorithm, with lower error values highlighted in bold format. We indicate the constellation as \texttt{Const.} for short. The data sequences indicate the number of \texttt{trial$\#$} in \texttt{Const.~$\#1\sim\#3$} and \texttt{trial$\#$.traj$\#$} in \texttt{Const.~$\#4$}, where \texttt{m$\#$} indicates a trial of data collected manually.}
  \begin{tabularx}{\linewidth}{p{0.02cm} p{0.05cm} c c c c }
  \toprule
    & & \multicolumn{2}{c}{TDOA2} &\multicolumn{2}{c}{TDOA3} \\
  \multicolumn{2}{c}{Data Seq.} & ESKF &Batch &ESKF &Batch \\
  \midrule
  \multirow{6}{*}{\rotatebox[origin=c]{90}{Const. 1}} & 1 & 0.108   &\textbf{0.097}   &0.144   &\textbf{0.099}  \\
      & 2 & 0.122      &\textbf{0.107}   &0.144      &\textbf{0.104}  \\
      & 3 & 0.109       &\textbf{0.097}   &0.123     &\textbf{0.095}  \\
      & 4 & 0.111       &\textbf{0.099}   &0.127     &\textbf{0.093}  \\
      & 5 & 0.115       &\textbf{0.106}   &0.149     &\textbf{0.120}  \\
      & 6 & 0.123       &\textbf{0.107}   &0.135     &\textbf{0.103}  \\
 \midrule
  \multirow{6}{*}{\rotatebox[origin=c]{90}{Const. 2}} & 1  & 0.103    &\textbf{0.098}     &0.114     &\textbf{0.082}  \\
      & 2   &0.114     &\textbf{0.096}   &0.116      &\textbf{0.078}  \\
      & 3   &\textbf{0.108}    &0.109   &0.117      &\textbf{0.079}  \\
      & 4 & 0.094      &\textbf{0.081}   &0.121    &\textbf{0.074}  \\
      & 5 & 0.122     &\textbf{0.099}   &0.125     &\textbf{0.076}  \\
      & 6 & 0.106      &\textbf{0.094}   &0.117     &\textbf{0.097}  \\
 \midrule
  \multirow{10}{*}{\rotatebox[origin=c]{90}{Const. 3}} & 1  & 0.245    &\textbf{0.127}     &0.232       &\textbf{0.144}  \\
      & 2 & 0.194     &\textbf{0.100}    &0.225      &\textbf{0.116}  \\
      & 3 & 0.189     &\textbf{0.093}    &0.214     &\textbf{0.124}  \\
      & 4 & 0.215    &\textbf{0.116}    &0.204     &\textbf{0.114}  \\
      & 5 & 0.220     &\textbf{0.117}    &0.241     &\textbf{0.126}  \\
      & 6 & 0.193     &\textbf{0.085}    &0.211      &\textbf{0.099}  \\
      & m1 &0.265    &\textbf{0.159}     &N/A     &N/A   \\
      & m2 &0.335     &\textbf{0.196}     &N/A     &N/A   \\
      & m3 & N/A    &N/A   &0.340      &\textbf{0.204}  \\
      & m4 & N/A   &N/A   &0.355       &\textbf{0.182}  \\
 \midrule
  \multirow{21}{*}{\rotatebox[origin=c]{90}{Const. 4}} & 1.1  & 0.201     &\textbf{0.176}     &0.207       &\textbf{0.146}  \\
      & 1.2 & 0.168   &\textbf{0.135}     &0.163       &\textbf{0.117}  \\
      & 1.3 & 0.194   &\textbf{0.146}   &0.190         &\textbf{0.139}  \\
      & 2.1 & 0.231       &\textbf{0.202}     &0.221    &\textbf{0.164}  \\
      & 2.2 & 0.200       &\textbf{0.158}   &0.219     &\textbf{0.170} \\
      & 2.3 & 0.222       &\textbf{0.185}   &0.238     &\textbf{0.177} \\
      & 3.1 & 0.919       &\textbf{0.788}   &0.742     &\textbf{0.563} \\
      & 3.2 & 0.718       &\textbf{0.624}   &0.558     &\textbf{0.482} \\
      & 3.3 & 0.651       &\textbf{0.571}   &0.674     &\textbf{0.529} \\
      & 4.1 & 0.818       &\textbf{0.704}   &0.759     &\textbf{0.612} \\
      & 4.2 & 0.766       &\textbf{0.648}   &0.845     &\textbf{0.669} \\
      & 4.3 & 0.735       &\textbf{0.636}   &0.838     &\textbf{0.669} \\
      & 5.1 & 0.473       &\textbf{0.397}   &0.432     &\textbf{0.334} \\
      & 5.2 & 0.470       &\textbf{0.397}   &0.435     &\textbf{0.312} \\
      & 5.3 &  0.395     &\textbf{0.340}   &0.443    &\textbf{0.325}  \\
      & 6.1 & 0.509      &\textbf{0.434}   &0.623     &\textbf{0.486} \\
      & 6.2 & 0.459       &\textbf{0.370}   &0.618      &\textbf{0.458} \\
      & 6.3 & 0.517       &\textbf{0.451}   &0.578     &\textbf{0.487} \\
      & m1  & 0.529       &\textbf{0.519}   &N/A       &N/A  \\
      & m2  & \textbf{0.587}       &0.613   &N/A      &N/A  \\
      & m3  & 0.333       &\textbf{0.266}   &N/A     &N/A  \\
  \bottomrule
  \end{tabularx}
  \label{tab:benchmark_rmse}
\end{table}
It is reported in the literature that the low-cost DW1000 UWB chips suffer from systematic measurement biases~\citep{zhao2021learning,ledergerber2017ultra}. Also, the UWB radio measurements are often corrupted with multi-path and NLOS signal propagation in real-world scenarios. In cluttered indoor environments, multi-path and NLOS radio propagation cannot be avoided in general. We summarized the UWB TDOA measurement $d_{23}$ in constellation $\#4$ in different LOS/NLOS scenarios in Figure~\ref{fig:flight-meas-comp}. The quadrotor was commanded to execute the same and repeated circle trajectory. We can observe in Figure~\ref{fig:flight-meas-comp}b-d that static obstacles consistently influence the UWB measurements. Also, UWB measurements can be completely blocked due to severe NLOS conditions. However, in dynamic NLOS scenarios (see Figure~\ref{fig:flight-meas-comp}e and f), the induced measurement errors do not remain consistent.  Hence, the flight dataset collected in constellation $\#4$ can be used to design new algorithms to cope with UWB measurement errors and noise so as to achieve robust and accurate UWB-based positioning performance in such challenging and highly dynamic environments.

In the past decade, researchers have dedicated efforts to enhance UWB TDOA localization performance while keeping costs low through the use of economical hardware. Despite these efforts, non-line-of-sight (NLOS) and multi-path radio propagation remain the major factors hindering the localization performance of UWB-based systems.
Identifying the measurement outliers and systematically handling the biased and non-Gaussian noise distributions~\citep{huang2022incremental,zhao2023uncertainty} in cluttered environments remains to be promising research directions. Additionally, the exploration of continuous-time estimation techniques~\citep{li2023continuous,goudar2023continuous} emerges as another promising research direction for asynchronous UWB-inertial localization systems. Furthermore, it is necessary to conduct observability analysis~\citep{goudar2021online} on system states and properly address the unobserved states to achieve consistent estimation~\citep{lisus2023know}.

\section{Conclusion}
\label{sec:conclusion}
In this paper, we present the UTIL dataset, a comprehensive UWB TDOA dataset based on the low-cost DWM1000 UWB modules.  Our dataset consists of \textit{(i)} an UWB identification dataset under various LOS and NLOS conditions and \textit{(ii)} a multimodal flight dataset collected with a cumulative total of around $150$ minutes of real-world flights in cluttered indoor environments with four anchor constellations. Obstacles of different types of materials commonly used in indoor settings, including cardboard, metal, wood, plastic, and foam, were used to create NLOS scenarios. During the flights, we collected raw UWB TDOA measurements with additional onboard sensor data (IMU, optical flow, and ToF laser) and millimeter-accurate ground truth data from a motion capture system onboard a quadrotor platform. The combination of multimodal onboard sensors, different anchor constellations, two TDOA modalities, and diverse cluttered scenarios contained in this dataset facilitates in-depth comparisons of UWB TDOA-based quadrotor localization capabilities. We hope this dataset can foster research in improving the UWB TDOA-based positioning performance in cluttered indoor environments. 
\section*{Acknowledgments}
\label{sec:acknowledgments}

We would like to thank Kristoffer Richardsson and Tobias Antonsson (Bitcraze) for their guidance on the software and hardware development of the customized quadrotor platform. This work was supported in part by the Natural Sciences and Engineering Research Council of Canada (NSERC) and in part by the Canada CIFAR AI Chairs Program.

% ack section
% \begin{acks}
% The acknowledge section.
% \end{acks}

%%Harvard (name/date)%\bibliographystyle{SageH}
\bibliographystyle{SageH}
\bibliography{reference.bib}

% \begin{thebibliography}{99}
% \bibitem[Kopka and Daly(2003)]{R1}
% Kopka~H and Daly~PW (2003) \textit{A Guide to \LaTeX}, 4th~edn.
% Addison-Wesley.

% \bibitem[Lamport(1994)]{R2}
% Lamport~L (1994) \textit{\LaTeX: a Document Preparation System},
% 2nd~edn. Addison-Wesley.

% \bibitem[Mittelbach and Goossens(2004)]{R3}
% Mittelbach~F and Goossens~M (2004) \textit{The \LaTeX\ Companion},
% 2nd~edn. Addison-Wesley.

% \end{thebibliography}

\end{document}